\documentclass{article}
\usepackage{ijcai22}

\usepackage{times}
\usepackage{soul}
\usepackage{url}
\usepackage[hidelinks]{hyperref}
\usepackage[utf8]{inputenc}
\usepackage[small]{caption}
\usepackage{graphicx}
\usepackage{amsmath}
\usepackage{amsthm}
\usepackage{booktabs}
\usepackage{algorithm}
\usepackage{algorithmic}
\usepackage{multirow}
\usepackage{url}
\usepackage{enumitem}
\usepackage{amsfonts}
\DeclareMathOperator*{\argmax}{argmax} 

\title{Speaker-Guided Encoder-Decoder Framework for Emotion Recognition in Conversation}

\author{
Yinan Bao$^{1,2}$\and
Qianwen Ma$^{1,2}$\and
Lingwei Wei$^{1,2}$\and
Wei Zhou$^{1,}$\thanks{Corresponding author.}\And
Songlin Hu$^{1,2}$
\affiliations
$^{1}$Institute of Information Engineering, Chinese Academy of Sciences \\
$^{2}$School of Cyber Security, University of Chinese Academy of Sciences
\emails
\{baoyinan, maqianwen, weilingwei, zhouwei, husonglin\}@iie.ac.cn}

\begin{document}

\maketitle

\begin{abstract}

The emotion recognition in conversation (ERC) task aims to predict the emotion label of an utterance in a conversation. Since the dependencies between speakers are complex and dynamic, which consist of intra- and inter-speaker dependencies, the modeling of speaker-specific information is a vital role in ERC. Although existing researchers have proposed various methods of speaker interaction modeling, they cannot explore dynamic intra- and inter-speaker dependencies jointly, leading to the insufficient comprehension of context and further hindering emotion prediction. To this end, we design a novel speaker modeling scheme that explores intra- and inter-speaker dependencies jointly in a dynamic manner. Besides, we propose a \textbf{S}peaker-\textbf{G}uided \textbf{E}ncoder-\textbf{D}ecoder (SGED) framework for ERC, which fully exploits speaker information for the decoding of emotion. We use different existing methods as the conversational context encoder of our framework, showing the high scalability and flexibility of the proposed framework. Experimental results demonstrate the superiority and effectiveness of SGED.

  
\end{abstract}


\section{Introduction}



Emotion recognition in conversation (ERC) aims to identify the emotion of each utterance in a conversation. Due to its potential applications in creating empathetic dialogue systems~\cite{Zhou:AAAI18}, social media analysis~\cite{Li:IJCAI19}, and so on, it has been concerned by a considerable number of researchers recently.

\begin{figure}[!htbp]
	\centering
	\includegraphics[scale=0.32]{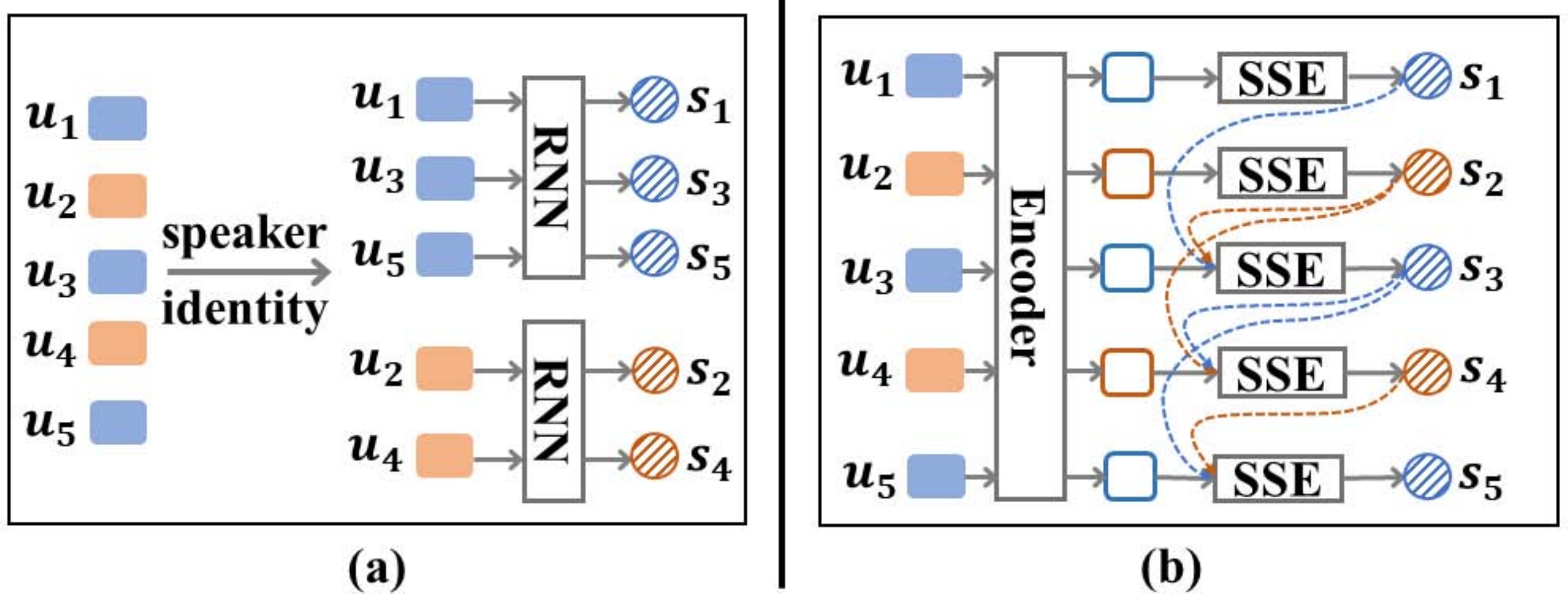}
	\vspace{-0.8\baselineskip}
	\caption{(a) Speaker modeling scheme of existing dynamic speaker-specific modeling methods, which is based on speaker identity to explore intra-speaker dependency. (b) Our novel speaker modeling scheme, which uses the speaker state encoder (SSE) to explore intra- and inter-speaker dependencies dynamically. $u_i$ means the $i_{th}$ utterance in a dialogue and $s_i$ means the corresponding speaker state vector of $u_i$.}
	\label{fig:intro}
	\vspace{-1.5\baselineskip}
\end{figure}

Unlike vanilla emotion recognition of sentences, emotions expressed in a conversation are dynamic and can be affected by many factors such as surrounding conversational context, and speaker dependencies~\cite{Poria:IEEE19}. Consequently, ERC models require a strong ability to model context, select crucial information, and capture speaker dependencies, which is more challenging. Among all the factors, speaker information is vital for tracking the emotional dynamics of conversations, which consists of two important aspects: intra- and inter-speaker dependencies~\cite{MELD:ACL19}. Intra-speaker dependency, also known as emotional inertia, deals with the aspect of emotional influence that speakers have on themselves during conversations~\cite{2010emotional}. In contrast, inter-speaker dependency refers to the emotional influences that the counterparts induce into a speaker~\cite{Survey:IEEE19}. Exploring intra-speaker dependency and inter-speaker dependency jointly is conducive to comprehending the complex dialogue context that implies various emotions.

Numerous approaches have been proposed for ERC with a focus on intra- and inter-speaker dependencies modeling. They can be divided into two categories: static speaker-specific modeling~\cite{ConGCN:IJCAI19,DialogueGCN:EMNLP19,HiTrans:COLING20,DialogXL:AAAI19} and dynamic speaker-specific modeling~\cite{DialogueRNN:AAAI19,DialogueCRN:ACL21}. For the static speaker-specific modeling methods, they utilize speaker information attaching to each utterance to specify different connections between utterances. In this way, the injected prior speaker information is static and shows the limited ability to facilitate conversational context modeling. Intuitively, the speaker state that is affected by the intra- and inter-speaker dependencies, changes dynamically during a dialogue. Different states of a certain speaker greatly affect various emotional expressions. Thus, it is necessary to implement dynamic speaker-specific modeling.


As shown in Figure~\ref{fig:intro}(a), existing dynamic speaker-specific modeling methods~\cite{DialogueRNN:AAAI19,DialogueCRN:ACL21} design speaker-specific recurrent neural networks (RNNs) to capture self-dependencies between adjacent utterances of the same speaker. In this scheme, utterances are classified based on speaker identity. However, sequences composed of separate utterances attaching to the same speaker are disjunctive and incomplete, which hinder the comprehension of the context. Moreover, overreliance on the speaker identity hinders the modeling of dynamic inter-speaker dependency. In this way, the model can't explore dynamic interactions between different speakers, which greatly affects the psychology of speakers, and then results in different emotional expressions. Nonetheless, how to explore intra- and inter-speaker dependencies that change dynamically is still unresolved.

To alleviate the problem, we design a novel speaker modeling scheme that explores intra- and inter-speaker dependencies dynamically. Besides, we propose a speaker-guided encoder-decoder framework (SGED), which can combine with various existing methods. The encoder module consists of two parts: conversational context encoder (CCE) and speaker state encoder (SSE). We use different existing methods which ignore dynamic speaker dependencies modeling as the CCE. For the SSE, we design a novel speaker modeling scheme to generate a dynamic speaker state sequence that depends on the speaker's previous state (intra-speaker dependency) and other neighboring speakers' states (inter-speaker dependency). For the decoder module based on RNN, we use speaker states to guide the decoding of utterances' emotions step by step. Extensive experiments indicate the effectiveness of SGED. In summary, our contributions are three-fold:


\begin{itemize}[topsep = 0 pt,leftmargin = *]
\setlength{\itemsep}{2pt}  
\setlength{\parsep}{0pt}  
\setlength{\parskip}{0pt}


\item To fully exploit the complex interactions between speakers that are vital for context comprehension, we propose a speaker-guided encoder-decoder framework (SGED) for ERC, formulating the modeling of speaker interactions as a flexible component.

\item We are the first to explore dynamic intra- and inter-speaker dependencies jointly through a subtly designed speaker state encoder, which explores complex interactions between speakers effectively.



\item Experimental results on three benchmark datasets demonstrate the effectiveness and high scalability of SGED.

\end{itemize}

\section{Related Work}

Recently, ERC has received increasing attention from the NLP community. Unlike traditional emotion recognition tasks, ERC models should be endowed with the ability of speaker interaction modeling and conversational context modeling.

\noindent\textbf{Speaker Interaction Modeling} According to whether the speaker-specific information is modeled dynamically during the dialogue, speaker modeling methods can be divided into two categories: static and dynamic speaker-specific modeling. For the static methods, some of them~\cite{DialogueGCN:EMNLP19,ConGCN:IJCAI19} treat dialogue as a graph and inject prior speaker information as different relations between utterances or treat speakers as nodes in the graph. DialogXL~\cite{DialogXL:AAAI19} changes the masking strategies of self-attention to capture intra- and inter-speaker dependencies. HiTrans~\cite{HiTrans:COLING20} exploits an auxiliary task to classify whether two utterances belong to the same speaker to make the model speaker-sensitive. However, these methods inject prior static speaker information into the model, neglecting that speaker states change dynamically according to the varying intra- and inter-speaker dependencies along with the dialogue. By contrast, DialogueRNN~\cite{DialogueRNN:AAAI19} and DialogueCRN~\cite{DialogueCRN:ACL21} utilize RNNs to update speakers' states in a dynamic manner. Nonetheless, they only consider intra-speaker dependency, ignoring inter-speaker dependency because of the limited speaker modeling scheme. Thus, we propose a speaker-guided encoder-decoder framework, which explores dynamic intra- and inter-speaker dependencies jointly.

\noindent\textbf{Conversational Context Modeling} According to the methods of conversational context modeling, existing models can be mainly divided into graph-based methods~\cite{DialogueGCN:EMNLP19,ConGCN:IJCAI19,DialogXL:AAAI19,KET:EMNLP19} and recurrence-based methods~\cite{DialogueRNN:AAAI19,DialogueCRN:ACL21,HiGRU:NAACL19,ICON:EMNLP18}. Recently, DAG-ERC~\cite{DAG-ERC:ACL21} is proposed to model the dialogue in the form of a directed acyclic graph, which allows the advantages of graph-based methods and recurrence-based methods to complement each other. Besides, some methods have exploited external commonsense
knowledge to enrich the context representation learning~\cite{KET:EMNLP19,COSMIC:EMNLP20}. We use different methods as the context encoder, proving the effectiveness and high scalability of our proposed speaker-guided encoder-decoder framework. 


\begin{figure*}[!htbp]
	\centering
	\includegraphics[scale=0.22]{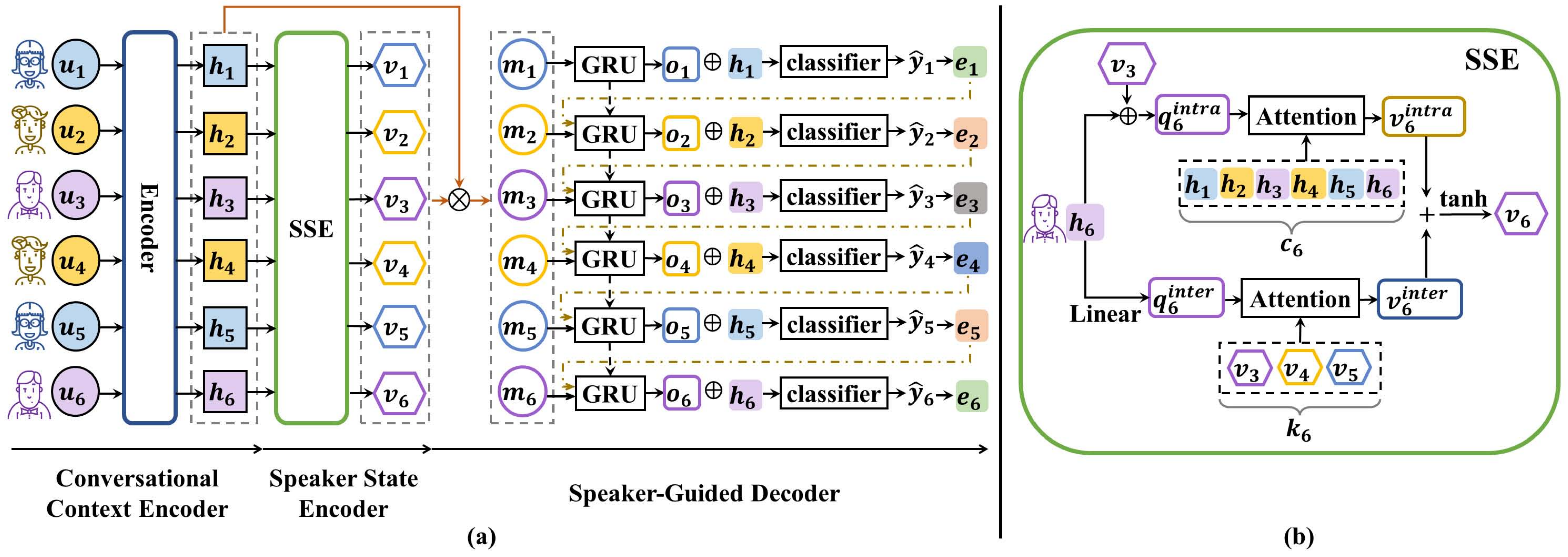}
	\vspace{-1.0\baselineskip}
	\caption{An overview of SGED. SSE represents the speaker state encoder. As shown in the right part of the figure, we take $u_6$ as an example to show how to obtain the speaker state vector of it.}
	\label{fig:model}
	\vspace{-1\baselineskip}
\end{figure*}

\section{Preliminary}

\subsection{Problem Definition}

A conversation containing $N$ utterances is denoted as $C = {\{u_1, u_2, ..., u_N}\}$. Each utterance $u_i$ consists of $n_i$ tokens, namely $u_i = {\{x_{i1}, x_{i2}, ..., x_{in_i}}\}$. There are $S$ speakers $s_1, s_2, ..., s_S (S\geq 2)$ in a dialogue. Utterance $u_i$ is uttered by speaker $s_{\phi(u_i)}$ where the function $\phi$ maps $u_i$ to its corresponding speaker. If $u_i$ isn't the first utterance spoken by $s_{\phi(u_i)}$ in the conversation, we denote $u_{\psi(u_i)}$ as the last previous utterance expressed by the same speaker where $0< \psi(u_i)<i$. We denote ${\{u_j | \forall{j}, \psi(u_i) \leq j < i}\}$ as the local information for $u_i$. ERC task aims to predict the emotion label $y_i$ for each utterance $u_i$ in the conversation $C$.

\subsection{Textual Feature Extraction}

By convention, we use the pre-trained language model to extract utterance-level representations. Following Shen et al~\shortcite{DAG-ERC:ACL21}, we fine-tune the pre-trained language model RoBERTa-Large~\cite{RoBERTa:arXiv19} on each ERC dataset first and then freeze its parameters while training our proposed framework. Concretely, we first append a special token [CLS] at the beginning of a given utterance $u_i$ to obtain the input sequence ${\{[CLS], x_{i1}, x_{i2}, ..., x_{in_i}}\}$. Afterward, we extract a representation of $u_i$ from the [CLS]'s embedding of the last layer with a dimension of 1,024.

\section{Methodology}


As illustrated in Figure~\ref{fig:model}, there are three components in our proposed framework SGED: (1) conversational context encoder; (2) speaker state encoder; (3) speaker-guided decoder.

\subsection{Conversational Context Encoder}



We take the existing baseline as the conversation context encoder of SGED, which is easily replaceable.\footnote{In the experiments, we take different existing methods, namely bc-LSTM+att~\cite{bcLSTM:ACL17}, DialogueGCN~\cite{DialogueGCN:EMNLP19}, and DAG-ERC~\cite{DAG-ERC:ACL21}, for the conversational context encoder.} For a given dialogue $C$, we generate the utterance representation vector sequence $\mathbf{H} = [\mathbf{h}_1, \mathbf{h}_2, ..., \mathbf{h}_N]$ as follows:\begin{equation} \label{equ:context_encoder}
\begin{split}
& \mathbf{H} = Encoder(C) \,, \\
\end{split}
\end{equation}
where $Encoder$ means the replaceable conversational context encoder.


\subsection{Speaker State Encoder}

Afterward, we generate a speaker state vector for each utterance based on the utterance representations from the conversational context encoder. The dynamic speaker state has two factors: intra-speaker dependency and inter-speaker dependency. We take the sequence of utterance representations as input and model the two kinds of dependencies simultaneously to obtain the final speaker state vectors sequentially.



\subsubsection{Intra-Speaker Dependency Modeling}

For a given utterance $u_i$ which is uttered by speaker $s_{\phi(u_i)}$, we take the last previous speaker state of $s_{\phi(u_i)}$ and previous context $\mathbf{c}_i = [\mathbf{h}_1, \mathbf{h}_2, ..., \mathbf{h}_i] \in \mathbb{R}^{h\times i}$ as input, modeling the intra-speaker dependency implied in the dialogue.

First, we utilize the last previous speaker state vector of $s_{\phi(u_i)}$ and the representation vector of $u_i$ to obtain the query vector $\mathbf{q}^{intra}_i$ as follows:\begin{equation} \label{equ:query_vector}
\begin{split}
& \mathbf{q}^{intra}_i = \mathbf{W}^{intra}_q [\mathbf{v}_{\psi(u_i)} || \mathbf{h}_i] + \mathbf{b}^{intra}_q \,, \\
\end{split}
\end{equation}
where $\mathbf{v}_{\psi(u_i)} \in \mathbb{R}^{h\times 1}$ means the last previous speaker state vector of $s_{\phi(u_i)}$; $||$ means the concatenation operation; $\mathbf{W}^{intra}_q\in \mathbb{R}^{h\times 2h}$ and $\mathbf{b}^{intra}_q\in \mathbb{R}^{1}$
are model parameters.

Then, we use $\mathbf{q}^{intra}_i$ as the query and previous context $\mathbf{c}_i$ as the key and value to implement the attention mechanism, generating the intra-speaker state vector of $u_i$ as follows:
\begin{equation}
\label{equ:intra_sp_state}
\begin{split}
& \alpha^{intra}_i = \mathbf{softmax}(\mathbf{W}_1 ( {\mathbf{q}^{intra}_i}\odot \mathbf{c}_i) + \mathbf{b}_1)  \,, \\
& \mathbf{v}^{intra}_i = \alpha^{intra}_i \mathbf{c}_i^T  \,, \\
\end{split}    
\end{equation}
where $\odot$ is an element-wise product operation; $\mathbf{W}_1 \in \mathbb{R}^{1\times h}$ and $\mathbf{b}_1\in \mathbb{R}^1$ are model parameters; $\alpha^{intra}_i\in \mathbb{R}^{1\times i}$ is the attention score to aggregate the previous context; $\mathbf{v}^{intra}_i\in \mathbb{R}^{1\times h}$ is the intra-speaker state vector of $u_i$.


\subsubsection{Inter-Speaker Dependency Modeling}

Simultaneously, we treat the representation vector of $u_i$ and the speaker state vectors of local information ${\{u_j|\forall{j}, \psi(u_i)\leq j < i}\}$ as input, modeling the latent inter-speaker dependency.

We adopt the following methods to generate the inter-speaker state vector of $u_i$:
\begin{equation}
\label{equ:inter_sp_state}
\begin{split}
& \mathbf{q}^{inter}_i = \mathbf{W}^{inter}_q \mathbf{h}_i + \mathbf{b}^{inter}_q  \,, \\
& \alpha^{inter}_i = \mathbf{softmax}(\mathbf{W}_2({\mathbf{q}^{inter}_i} \odot \mathbf{k}_i) + \mathbf{b}_2)  \,, \\
& \mathbf{v}^{inter}_i = \alpha^{inter}_i \mathbf{k}_i^T  \,, \\
\end{split}
\end{equation}
where $\mathbf{k}_i = {\{\mathbf{v}_j | \forall{j}, \psi(u_i)\leq j <i}\} \in \mathbb{R}^{h\times l}$ denotes the speaker state vectors of previous local information; $\odot$ means the element-wise product operation; $\mathbf{W}^{inter}_q\in \mathbb{R}^{h\times h}$, $\mathbf{b}^{inter}_q\in \mathbb{R}^{1}$, $\mathbf{W}_2\in \mathbb{R}^{1\times h}$, and $\mathbf{b}_2\in \mathbb{R}^1$ are model parameters; $\mathbf{v}^{inter}_i\in \mathbb{R}^{1\times h}$ is the inter-speaker state vector of $u_i$.

Finally, we fuse the intra-speaker state vector $\mathbf{v}^{intra}_i$ and the inter-speaker state vector $\mathbf{v}^{inter}_i$ to generate the final speaker state vector of $u_i$:
\begin{equation}
\label{equ:sp_state_fusion}
\begin{split}
& \mathbf{v}_i = \mathbf{tanh}(\mathbf{v}^{intra}_i + \mathbf{v}^{inter}_i)  \,, \\
\end{split}
\end{equation}
where $\mathbf{v}_i\in \mathbb{R}^{1\times h}$ is the speaker state vector of $u_i$. Note that if $u_i$ is the first utterance expressed by speaker $s_{\phi(u_i)}$ in the dialogue, we let $\mathbf{v}_i = \mathbf{h}_i$. For the given dialogue $C$, we obtain the speaker state vector sequence $\mathbf{V} = {\{\mathbf{v}_1, \mathbf{v}_2, ..., \mathbf{v}_N}\}$, which encodes speaker-specific intra-speaker dependency and inter-speaker dependency dynamically.

\subsection{Speaker-Guided Decoder}

To fully utilize the vital speaker-specific information implied in the conversation, we use the speaker state vector $\mathbf{v}_i$ to guide the decoding of $u_i$'s emotion. In addition, inherent relations between emotion tags also facilitate ERC~\cite{CESTa:SigDial20}. Therefore, we also use the embedding of the predicted emotion of $u_{i-1}$ to instruct the decoding of $u_i$'s emotion. We subtly design a speaker-guided emotion decoder based on (gated recurrent unit) GRU to decode each utterance's emotion in a given dialogue sequentially. 

Given an utterance $u_i$, we first match the speaker state vector $\mathbf{v}_i$ with the utterance representation vector $\mathbf{h}_i$ as follows:\begin{equation}
\label{equ:decoder_match}
\begin{split}
& \mathbf{m}_i = \mathbf{ReLU}(\mathbf{v}_i \odot (\mathbf{W}_m \mathbf{h}_i + \mathbf{b}_m)^T)  \,, \\
\end{split}
\end{equation}
where $\mathbf{W}_m\in \mathbb{R}^{h\times h}$ and $\mathbf{b}_m\in \mathbb{R}^{1}$ are trainable parameters for feature transformation; $\mathbf{m}_i\in \mathbb{R}^{1\times h}$ is the generated match vector.

Afterward, we concatenate the match vector $\mathbf{m}_i$ and the embedding $\mathbf{e}_{i-1}$ of the predicted emotion of $u_{i-1}$ and then feed it to GRU:\begin{equation}
\label{equ:decoder}
\begin{split}
& \mathbf{o}_i = \overrightarrow{\mathbf{GRU}}([\mathbf{m}_i || \mathbf{e}_{i-1}], \mathbf{o}_{i-1})  \,. \\
\end{split}
\end{equation}

Finally, we take the concatenation of $\mathbf{h}_i\in  \mathbb{R}^{h\times 1}$ and $\mathbf{o}_i\in \mathbb{R}^{1\times h}$ as the final representation of $u_i$, and feed it to a feed-forward neural network for emotion prediction:\begin{equation}
\label{equ:prediction}
\begin{split}
& \mathbf{z}_i = \mathbf{ReLU}(\mathbf{W}_o [\mathbf{h}_i || \mathbf{o}_i^T] + \mathbf{b}_o)  \,, \\
& \mathcal{P}_i = \mathbf{softmax}(\mathbf{W}_z \mathbf{z}_i + \mathbf{b}_z)  \,, \\
& \hat{y}_i = \argmax_g \mathcal{P}_i[g]  \,, \\
& \mathbf{e}_i = \mathbf{E}[\hat{y}_i]  \,, \\
\end{split}
\end{equation}
where $\mathbf{W}_o\in \mathbb{R}^{h\times 2h}$, $\mathbf{W}_z\in \mathbb{R}^{L\times h}$, $\mathbf{b}_o, \mathbf{b}_z\in \mathbb{R}^1$ are trainable parameters; $L$ is the class number of the dataset; $\hat{y}_i$ is the predicted label of $u_i$; $\mathbf{E}$ is the look-up table of emotion embeddings which is initialized randomly at the beginning; $\mathbf{e}_i$ is the embedding of the predicted emotion of $u_i$.

To train the model, we use cross-entropy loss function:\begin{equation}
\label{equ:loss_function}
\begin{split}
& \mathcal{L}(\theta) = -\sum_{j=1}^M \sum_{t=1}^{N_j} log\mathcal{P}_{j,t}[y_{j,t}]  \,, \\
\end{split}
\end{equation}
where $M$ is the number of dialogues in the training set; $N_j$ is the number of utterances in the $j$-th dialogue; $y_{j,t}$ is the ground-truth label; $\theta$ is the collection of trainable parameters.


\section{Experimental Settings}

\subsection{Datasets}
We evaluate our proposed framework on three ERC benchmark datasets. The statistics of them are shown in Table~\ref{tab:data_statistic}.

\noindent\textbf{IEMOCAP}~\cite{IEMOCAP:08}: A two-party conversation dataset for ERC, which includes 6 types of emotion, that is \textit{neutral}, \textit{happiness}, \textit{sadness}, \textit{anger}, \textit{frustrated}, and \textit{excited}. Since this dataset has no validation set, we follow~\cite{DAG-ERC:ACL21} to use the last 20 conversations in the training set for validation.

\noindent\textbf{MELD}~\cite{MELD:ACL19}: A multi-party conversation dataset collected from the TV series \textit{Friends}. The emotion labels are: \textit{neutral}, \textit{happiness}, \textit{surprise}, \textit{sadness}, \textit{anger}, \textit{disgust}, and \textit{fear}.

\noindent\textbf{EmoryNLP}~\cite{EmoryNLP:18}: A multi-party conversation dateset collected from \textit{Friends}, but varies from MELD in the choice of scenes and emotion labels. There are 7 types of emotion, namely \textit{neutral}, \textit{sad}, \textit{mad}, \textit{scared}, \textit{powerful}, \textit{peaceful}, and \textit{joyful}.

Although all of the datasets contain multimodal information, we only focus on the textual information for experiments. Following existing work~\cite{DAG-ERC:ACL21}, we choose weighted-average F1 for evaluation metrics.

\begin{table}\small
\centering
\renewcommand\arraystretch{0.8}
\setlength{\tabcolsep}{7pt}{
\begin{tabular}{l|c|c|c|c|c|c}
\toprule
    \multirow{2}{*}{Dataset} & 
    \multicolumn{3}{|c|}{Conversations} &
    \multicolumn{3}{|c}{Utterances} \\
        ~ & Train & Val & Test & Train & Val & Test \\
    \midrule[0.7pt]
    
    IEMOCAP & \multicolumn{2}{c|}{120} & 31 & \multicolumn{2}{c|}{5810} & 1623 \\
    MELD & 1038 & 114 & 280 & 9989 & 1109 & 2610 \\
    EmoryNLP & 713 & 99 & 85 & 9934 & 1344 & 1328 \\
    
    \bottomrule
\end{tabular}
}
\caption{The statistics of three datasets.}
\vspace{-1\baselineskip}
\label{tab:data_statistic}
\end{table}

\subsection{Compared Methods}

We compare our method with the following baselines. 

\textbf{KET}~\cite{KET:EMNLP19} is a knowledge-enriched transformer that extracts commonsense knowledge based on a graph attention mechanism. \textbf{COSMIC}~\cite{COSMIC:EMNLP20} incorporates different elements of commonsense and leverages them to learn self-speaker dependency. \textbf{DialogueRNN}~\cite{DialogueRNN:AAAI19} uses speaker-specific GRU to model intra-speaker dependency and generate speaker states sequentially in a dynamic manner. Another dynamic speaker-specific modeling method is \textbf{DialogueCRN}~\cite{DialogueCRN:ACL21}, which utilizes speaker-level context to model dynamic intra-speaker dependency and designs a multi-turn reasoning module to imitate human cognitive thinking. \textbf{bc-LSTM+att}~\cite{bcLSTM:ACL17} is a contextual LSTM network without speaker information modeling, which is based on an attention mechanism to re-weight the features. \textbf{DialogueGCN}~\cite{DialogueGCN:EMNLP19} is a graph-based method that treats speaker dependencies as static information and designs different relations between utterances. \textbf{DAG-ERC}~\cite{DAG-ERC:ACL21} models a dialogue in the form of a directed acyclic graph and utilizes speaker identity to establish connections between utterances.

We use the same feature extractor \textbf{RoBERTa}~\cite{RoBERTa:arXiv19} as provided by Shen et al.,~\shortcite{DAG-ERC:ACL21}. We replace the textual feature in the previous methods with our extracted features first. Then, we use previous methods without dynamic speaker-specific information modeling (RoBERTa, bc-LSTM+att, DialogueGCN, DAG-ERC\footnote{Note that we use the one-layer DAG-ERC as the context encoder, making the parameters fewer than the original DAG-ERC with multiple layers. In this way, the potential impact of increasing parameters provided by the decoder is eliminated. Our method with fewer parameters even surpasses the original DAG-ERC with more parameters, proving the effectiveness of our method more convincingly.}) as the conversational context encoder of the proposed speaker-guided encoder-decoder framework, investigating the effectiveness of our speaker modeling scheme.




\subsection{Implementation Details}

We utilize the validation set to tune parameters on each dataset and adopt Adam~\cite{Adam:ICLR15} as the optimizer. The learning rate and batch size are selected from \{1e-5, 5e-5, 1e-4, 5e-4\} and \{8, 16, 32\}, respectively. The dimension of hidden vector $h$ is set to 300 and the dimension of label embedding is set to 100. The feature size for the RoBERTa extractor is 1,024. We use the fine-tuned RoBERTa features provided by Shen et al.,~\shortcite{DAG-ERC:ACL21}. Each training process contains 60 epochs. The results of our proposed SGED framework reported in the experiments are all based on the average score of 5 random runs on the test set.


\begin{table}
    \centering
    \renewcommand\arraystretch{0.9}
    \setlength{\tabcolsep}{3.3pt}{
    \begin{tabular}{l|c|c|c}
        \toprule
        Model & MELD & EmoryNLP & IEMOCAP \\
        \midrule[0.7pt]
        
        KET & 58.18 & 33.95 & 59.56 \\
        
        COSMIC & 65.21 & 38.11 & 65.28 \\
        
        
        \midrule[0.7pt]
        
        DialogueRNN & 57.03 & - & 62.75 \\
        \quad + RoBERTa & 63.61 & 37.44 & 64.76  \\
        
        \midrule[0.3pt]
        
        DialogueCRN & 58.39 & - & 66.20  \\
        \quad + RoBERTa & 63.42 & 38.91 & 66.46  \\
        
        \midrule[0.7pt]
        
        RoBERTa & 62.88 & 37.78 & 63.38 \\
        \textbf{SGED + RoBERTa} & \textbf{63.34} & \textbf{38.47} & \textbf{64.11} \\
        
        \midrule[0.3pt]
        
        
        bc-LSTM+att & - & - & - \\
        \quad + RoBERTa & 62.95 & 38.28 & 64.51 \\
        \textbf{SGED + bc-LSTM+att} & \textbf{63.37} & \textbf{38.89} & \textbf{65.03} \\
        
        \midrule[0.3pt]
        
        DialogueGCN &  58.10 & - & 64.18 \\
        \quad + RoBERTa & 63.02 & 38.10 & 64.91 \\
        \textbf{SGED + DialogueGCN} & \textbf{64.55} & \textbf{39.73} & \textbf{65.90} \\
        
        \midrule[0.3pt]
        
        
        
        DAG-ERC & 63.65 & 39.02 & 68.03 \\
        DAG-ERC* & 63.39 & 38.84 & 67.45 \\
        \textbf{SGED + DAG-ERC*} & \textbf{65.46} & \textbf{40.24} & \textbf{68.53} \\
        
        
        
        
        
        
        
        
        
        \bottomrule
    \end{tabular}
    }
    \caption{Overall performance on the three datasets. We choose weighted-average F1 to evaluate each method. DAG-ERC* means that we use the one-layer DAG-ERC.}
    \label{tab:results}
\end{table}

\begin{table}
    \centering
    \renewcommand\arraystretch{0.9}
    \setlength{\tabcolsep}{6pt}{
    \begin{tabular}{l|l|l}
        \toprule
        Method & EmoryNLP & IEMOCAP  \\
        
        \midrule[0.7pt]
        
        \textbf{SGED + DAG-ERC*} & \textbf{40.24} & \textbf{68.53} \\
        
        w/o \textbf{SSE} - intra-speaker & 39.87 ($\downarrow$0.37) & 68.05 ($\downarrow$0.48) \\
        w/o \textbf{SSE} - inter-speaker & 39.86 ($\downarrow$0.38) & 67.72 ($\downarrow$0.81) \\
        w/o \textbf{SSE} & 39.17 ($\downarrow$1.07) & 67.67 ($\downarrow$0.86) \\
        
        
        w/o \textbf{SGD} & 38.96 ($\downarrow$1.28) & 67.30 ($\downarrow$1.23) \\
        
        
        
        w/o \textbf{SSE + SGD} & 38.84 ($\downarrow$1.40) & 67.45 ($\downarrow$1.08) \\
        
        
        \bottomrule
    \end{tabular}
    }
    \caption{Results of ablation study on EmoryNLP (multi-party) and IEMOCAP (two-party) dataset. We use the one-layer DAG-ERC as the conversational context encoder of our framework.}
    \label{tab:ablation_study}
    \vspace{-1\baselineskip}
\end{table}


\section{Results and Analysis}

\subsection{Overall Results}

Overall results of our proposed method and the baselines are reported on Table~\ref{tab:results}. The results demonstrate that methods combined with our speaker-guided encoder-decoder framework achieve significant improvement. 

As shown in Table~\ref{tab:results}, based on the same feature extractor RoBERTa, the framework proposed by us outperforms the other dynamic speaker-specific information modeling methods (\textbf{DialogueRNN}, \textbf{DialogueCRN}) in general. Moreover, \textbf{SGED + RoBERTa} which encodes each utterance independently in the context encoder, achieves comparable results with \textbf{DialogueRNN+RoBERTa}, proving the effectiveness of the SGED proposed by us. Taking different types of existing methods as the conversational context encoder, SGED which combines the novel speaker modeling scheme achieves significant improvement all the time. It indicates the high scalability and flexibility of SGED. Moreover, \textbf{SGED + DAG-ERC*} reaches the new state-of-the-art on the three benchmark datasets, even surpasses \textbf{COSMIC} which utilizes external commonsense knowledge.

\textit{\textbf{MELD and EmoryNLP:}} On these multi-party conversation datasets, we observe that our SGED framework makes varying degrees of promotion based on different conversational context encoders. For the recurrence-based method \textbf{bc-LSTM+att} and context-independent method \textbf{RoBERTa}, the promotion brought by the SGED framework is about \textbf{0.55\%} on average. However, for the graph-based methods \textbf{DialogueGCN} and \textbf{DAG-ERC}, our SGED framework achieves an increase of \textbf{1.66\%} on average. It is because the speaker state encoder of SGED is established based on the conversational context encoder. Thus, the context comprehension ability of the conversational context encoder affects the subsequent dynamic speaker-specific information modeling, and finally influences the decoding of emotion. The stable improvement brought by SGED indicates that the proposed dynamic speaker modeling scheme helps to explore crucial dependencies between speakers and benefits ERC.

\textit{\textbf{IEMOCAP:}} On the two-party conversation dataset, we observe that the improvement brought by SGED has little correlation with the type of context encoder, which is different from the results on the multi-party conversation datasets. Besides, the improvement of SGED on IEMOCAP is \textbf{0.83\%} on average, which is slightly lower than the improvements on MELD and EmoryNLP. We argue that the dependencies between speakers in multi-party conversations are more complex and intractable. Therefore, it is more necessary to exploit SGED to model speaker-specific states dynamically for multi-party conversations. 

\begin{figure*}[!htbp]
	\centering
	\includegraphics[scale=0.27]{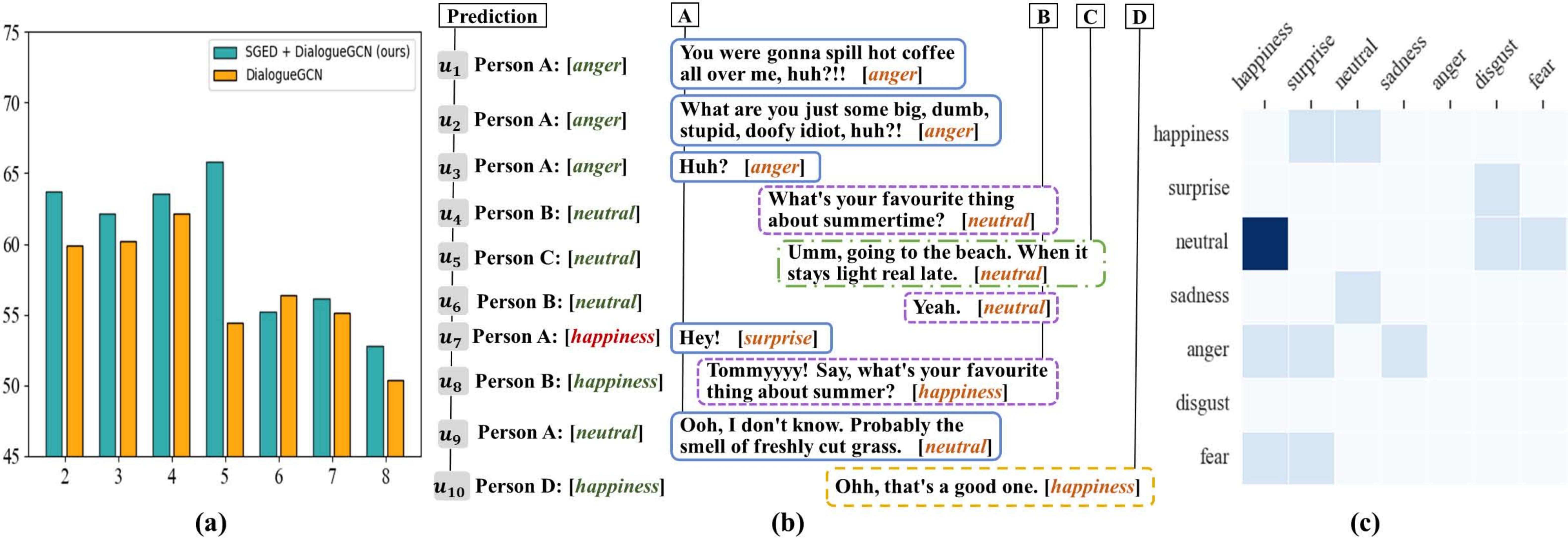}
	\vspace{-0.8\baselineskip}
	\caption{(a) The weighted-average F1 score of conversations with the different number of speakers on MELD, achieving by DialogueGCN and our SGED framework which takes DialogueGCN as the conversational context encoder. (b) An abridged dialogue from MELD in which our SGED framework recognizes most of the emotions correctly. (c) The confusion matrix of conversations with 6 speakers excluding the diagonal on MELD, is achieved by our SGED framework which takes DialogueGCN as the conversational context encoder.}
	\label{fig:case_study}
	\vspace{-1\baselineskip}
\end{figure*}

\subsection{Ablation Study}

To investigate the effects of different modules in SGED, we conduct several ablation studies on EmoryNLP and IEMOCAP. We remove 1) intra-speaker dependency in speaker state encoder (SSE), 2) inter-speaker dependency in SSE, 3) SSE, 4) speaker-guided decoder (SGD), and 5) SSE and SGD, respectively. The results are reported on Table~\ref{tab:ablation_study}.

\paragraph{\textbf{Effect of Different Dynamic Dependencies between Speakers.}} As shown in Table~\ref{tab:ablation_study}, the result on EmoryNLP declines when removing either intra-speaker dependency or inter-speaker dependency in the SSE. However, SGED without inter-speaker dependency performs worse than without intra-speaker dependency on IEMOCAP. It proves that both dynamic intra- and inter-speaker dependencies are important for multi-party conversations while dynamic inter-speaker dependency is more important for two-party conversations. Moreover, the result drops sharply when eliminating the entire speaker state encoder (SSE) that explores both intra- and inter-speaker dependencies in a dynamic manner, which indicates that the two kinds of dynamic dependency are complementary to each other.

\paragraph{\textbf{Effect of Speaker-Guided Decoder.}} To evaluate the effectiveness of the speaker-guided decoder (SGD), we remove it and only use the speaker state vector for prediction. As shown in Table~\ref{tab:ablation_study}, it is clear that the speaker-guided decoder proposed by us boosts model performance a lot. Removing the SGD not only has a great impact on multi-party conversations (\textbf{1.28\%}) but also two-party conversations (\textbf{1.23\%}), which proves that the aggregation of speaker state feature and semantic feature benefits emotion recognition in conversation.



\paragraph{\textbf{Effect of the Novel Speaker Modeling Scheme.}} Our proposed speaker modeling scheme consists of speaker state encoder (SSE) and speaker-guided decoder (SGD). As reported on Tabel~\ref{tab:ablation_study}, removing SSE and SGD simultaneously results in a significant drop. It indicates the effectiveness of the proposed speaker modeling scheme, which can capture dynamic speaker interactions and promote emotion prediction.




\subsection{Discussion}

As shown in Figure~\ref{fig:case_study} (a), we illustrate the results of dialogues with the different number of speakers on MELD, achieved by DialogueGCN and our SGED framework which takes DialogueGCN as the conversational context encoder. We observe that our proposed method exceeds DialogueGCN on dialogues with different speaker numbers except 6. Compared with DialogueGCN, our method achieves the greatest promotion on dialogues with 5 speakers. We will take an example with 5 speakers for the case study and conduct an error analysis on dialogues with 6 speakers.

\subsubsection{Case Study}
In Figure~\ref{fig:case_study} (b), we present an abridged dialogue that has 5 speakers originally and our method recognizes most emotions correctly. For person \texttt{A}, we recognize $u_3$ as \textit{anger} correctly though it is a short utterance that is intractable. We argue that the intra-speaker dependency of \texttt{A} captured by our method facilitates the recognition. Moreover, although person \texttt{A}'s previous utterances ($u_1, u_2, u_3$) all express \textit{anger}, our method recognizes $u_7$ as \textit{happiness} which is obviously different from \textit{anger}, based on the inter-speaker dependencies among person \texttt{C} ($u_5$), \texttt{B} ($u_6$), and \texttt{A} ($u_7$). For person \texttt{B}, we recognize $u_8$ as happiness correctly, due to the trade-off of inter-speaker dependency between person \texttt{A} ($u_7$) and \texttt{B} ($u_8$) and intra-speaker dependency of \texttt{B} captured by our method. It indicates that SGED models dynamic speaker states effectively with integration of intra-speaker dependency and inter-speaker dependency.



\subsubsection{Error Analysis}
In Figure~\ref{fig:case_study} (c), we present the confusion matrix of conversations with 6 speakers excluding the diagonal, achieved by our method. From the heat map, \textit{neutral} samples are more confused with non-neutral (e.g., \textit{happiness}) samples, which can be alleviated by adding cues from other modalities. In addition to this, distinguishing similar emotions (e.g., \textit{happiness} and \textit{surprise}) still disturbs our method.

\section{Conclusion}

In this paper, we propose a speaker-guided encoder-decoder framework (SGED) for ERC, which captures dynamic intra- and inter-speaker dependencies to benefit the understanding of the complex context. In this framework, we propose a novel speaker state encoder (SSE) and design a speaker-guided decoder (SGD). SSE is competent at exploring intra- and inter-speaker dependencies dynamically. Moreover, we treat different existing methods as the conversational context encoder of SGED, showing its high scalability and flexibility. Experiments on three benchmark datasets prove the superiority and effectiveness of our method. Especially, our method achieves a more significant improvement on the multi-party conversation datasets, indicating its strong ability to model dynamic multi-speaker states during a dialogue.


\bibliographystyle{named}
\bibliography{ijcai22}

\begin{thebibliography}{}

\bibitem[\protect\citeauthoryear{Busso \bgroup \em et al.\egroup
  }{2008}]{IEMOCAP:08}
Carlos Busso, Murtaza Bulut, Chi{-}Chun Lee, Abe Kazemzadeh, Emily Mower,
  Samuel Kim, Jeannette~N. Chang, Sungbok Lee, and Shrikanth~S. Narayanan.
\newblock {IEMOCAP:} interactive emotional dyadic motion capture database.
\newblock {\em Lang. Resour. Evaluation}, 42(4):335--359, 2008.

\bibitem[\protect\citeauthoryear{Ghosal \bgroup \em et al.\egroup
  }{2019}]{DialogueGCN:EMNLP19}
Deepanway Ghosal, Navonil Majumder, Soujanya Poria, Niyati Chhaya, and
  Alexander~F. Gelbukh.
\newblock Dialoguegcn: {A} graph convolutional neural network for emotion
  recognition in conversation.
\newblock In {\em {EMNLP-IJCNLP}}, pages 154--164, 2019.

\bibitem[\protect\citeauthoryear{Ghosal \bgroup \em et al.\egroup
  }{2020}]{COSMIC:EMNLP20}
Deepanway Ghosal, Navonil Majumder, Alexander~F. Gelbukh, Rada Mihalcea, and
  Soujanya Poria.
\newblock {COSMIC:} commonsense knowledge for emotion identification in
  conversations.
\newblock In Trevor Cohn, Yulan He, and Yang Liu, editors, {\em {EMNLP}},
  volume {EMNLP} 2020 of {\em Findings of {ACL}}, pages 2470--2481, 2020.

\bibitem[\protect\citeauthoryear{Hazarika \bgroup \em et al.\egroup
  }{2018}]{ICON:EMNLP18}
Devamanyu Hazarika, Soujanya Poria, Rada Mihalcea, Erik Cambria, and Roger
  Zimmermann.
\newblock {ICON:} interactive conversational memory network for multimodal
  emotion detection.
\newblock In {\em EMNLP}, pages 2594--2604, 2018.

\bibitem[\protect\citeauthoryear{Hu \bgroup \em et al.\egroup
  }{2021}]{DialogueCRN:ACL21}
Dou Hu, Lingwei Wei, and Xiaoyong Huai.
\newblock Dialoguecrn: Contextual reasoning networks for emotion recognition in
  conversations.
\newblock In {\em {ACL/IJCNLP}}, pages 7042--7052, 2021.

\bibitem[\protect\citeauthoryear{Jiao \bgroup \em et al.\egroup
  }{2019}]{HiGRU:NAACL19}
Wenxiang Jiao, Haiqin Yang, Irwin King, and Michael~R. Lyu.
\newblock Higru: Hierarchical gated recurrent units for utterance-level emotion
  recognition.
\newblock In {\em {NAACL-HLT}}, pages 397--406, 2019.

\bibitem[\protect\citeauthoryear{Kingma and Ba}{2015}]{Adam:ICLR15}
Diederik~P. Kingma and Jimmy Ba.
\newblock Adam: {A} method for stochastic optimization.
\newblock In {\em {ICLR}}, 2015.

\bibitem[\protect\citeauthoryear{Kuppens \bgroup \em et al.\egroup
  }{2010}]{2010emotional}
Peter Kuppens, Nicholas~B Allen, and Lisa~B Sheeber.
\newblock Emotional inertia and psychological maladjustment.
\newblock {\em Psychological science}, 21(7):984--991, 2010.

\bibitem[\protect\citeauthoryear{Li \bgroup \em et al.\egroup
  }{2019}]{Li:IJCAI19}
Runnan Li, Zhiyong Wu, Jia Jia, Yaohua Bu, Sheng Zhao, and Helen Meng.
\newblock Towards discriminative representation learning for speech emotion
  recognition.
\newblock In {\em {IJCAI}}, pages 5060--5066, 2019.

\bibitem[\protect\citeauthoryear{Li \bgroup \em et al.\egroup
  }{2020}]{HiTrans:COLING20}
Jingye Li, Donghong Ji, Fei Li, Meishan Zhang, and Yijiang Liu.
\newblock Hitrans: {A} transformer-based context- and speaker-sensitive model
  for emotion detection in conversations.
\newblock In {\em {COLING}}, pages 4190--4200, 2020.

\bibitem[\protect\citeauthoryear{Liu \bgroup \em et al.\egroup
  }{2019}]{RoBERTa:arXiv19}
Yinhan Liu, Myle Ott, Naman Goyal, Jingfei Du, Mandar Joshi, Danqi Chen, Omer
  Levy, Mike Lewis, Luke Zettlemoyer, and Veselin Stoyanov.
\newblock Roberta: {A} robustly optimized {BERT} pretraining approach.
\newblock {\em CoRR}, abs/1907.11692, 2019.

\bibitem[\protect\citeauthoryear{Majumder \bgroup \em et al.\egroup
  }{2019}]{DialogueRNN:AAAI19}
Navonil Majumder, Soujanya Poria, Devamanyu Hazarika, Rada Mihalcea,
  Alexander~F. Gelbukh, and Erik Cambria.
\newblock Dialoguernn: An attentive {RNN} for emotion detection in
  conversations.
\newblock In {\em {AAAI}}, pages 6818--6825, 2019.

\bibitem[\protect\citeauthoryear{Poria \bgroup \em et al.\egroup
  }{2017}]{bcLSTM:ACL17}
Soujanya Poria, Erik Cambria, Devamanyu Hazarika, Navonil Majumder, Amir Zadeh,
  and Louis{-}Philippe Morency.
\newblock Context-dependent sentiment analysis in user-generated videos.
\newblock In {\em {ACL}}, pages 873--883, 2017.

\bibitem[\protect\citeauthoryear{Poria \bgroup \em et al.\egroup
  }{2019a}]{MELD:ACL19}
Soujanya Poria, Devamanyu Hazarika, Navonil Majumder, Gautam Naik, Erik
  Cambria, and Rada Mihalcea.
\newblock {MELD:} {A} multimodal multi-party dataset for emotion recognition in
  conversations.
\newblock In {\em {ACL}}, pages 527--536, 2019.

\bibitem[\protect\citeauthoryear{Poria \bgroup \em et al.\egroup
  }{2019b}]{Poria:IEEE19}
Soujanya Poria, Navonil Majumder, Rada Mihalcea, and Eduard~H. Hovy.
\newblock Emotion recognition in conversation: Research challenges, datasets,
  and recent advances.
\newblock {\em {IEEE} Access}, 7:100943--100953, 2019.

\bibitem[\protect\citeauthoryear{Poria \bgroup \em et al.\egroup
  }{2019c}]{Survey:IEEE19}
Soujanya Poria, Navonil Majumder, Rada Mihalcea, and Eduard~H. Hovy.
\newblock Emotion recognition in conversation: Research challenges, datasets,
  and recent advances.
\newblock {\em {IEEE} Access}, 7:100943--100953, 2019.

\bibitem[\protect\citeauthoryear{Shen \bgroup \em et al.\egroup
  }{2021a}]{DialogXL:AAAI19}
Weizhou Shen, Junqing Chen, Xiaojun Quan, and Zhixian Xie.
\newblock Dialogxl: All-in-one xlnet for multi-party conversation emotion
  recognition.
\newblock In {\em {AAAI}}, pages 13789--13797, 2021.

\bibitem[\protect\citeauthoryear{Shen \bgroup \em et al.\egroup
  }{2021b}]{DAG-ERC:ACL21}
Weizhou Shen, Siyue Wu, Yunyi Yang, and Xiaojun Quan.
\newblock Directed acyclic graph network for conversational emotion
  recognition.
\newblock In {\em {ACL/IJCNLP}}, pages 1551--1560, 2021.

\bibitem[\protect\citeauthoryear{Wang \bgroup \em et al.\egroup
  }{2020}]{CESTa:SigDial20}
Yan Wang, Jiayu Zhang, Jun Ma, Shaojun Wang, and Jing Xiao.
\newblock Contextualized emotion recognition in conversation as sequence
  tagging.
\newblock In {\em SIGdial}, pages 186--195, 2020.

\bibitem[\protect\citeauthoryear{Zahiri and Choi}{2018}]{EmoryNLP:18}
Sayyed~M. Zahiri and Jinho~D. Choi.
\newblock Emotion detection on {TV} show transcripts with sequence-based
  convolutional neural networks.
\newblock In {\em {AAAI}}, volume {WS-18} of {\em {AAAI} Workshops}, pages
  44--52, 2018.

\bibitem[\protect\citeauthoryear{Zhang \bgroup \em et al.\egroup
  }{2019}]{ConGCN:IJCAI19}
Dong Zhang, Liangqing Wu, Changlong Sun, Shoushan Li, Qiaoming Zhu, and Guodong
  Zhou.
\newblock Modeling both context- and speaker-sensitive dependence for emotion
  detection in multi-speaker conversations.
\newblock In {\em {IJCAI}}, pages 5415--5421, 2019.

\bibitem[\protect\citeauthoryear{Zhong \bgroup \em et al.\egroup
  }{2019}]{KET:EMNLP19}
Peixiang Zhong, Di~Wang, and Chunyan Miao.
\newblock Knowledge-enriched transformer for emotion detection in textual
  conversations.
\newblock In {\em {EMNLP-IJCNLP}}, pages 165--176, 2019.

\bibitem[\protect\citeauthoryear{Zhou \bgroup \em et al.\egroup
  }{2018}]{Zhou:AAAI18}
Hao Zhou, Minlie Huang, Tianyang Zhang, Xiaoyan Zhu, and Bing Liu.
\newblock Emotional chatting machine: Emotional conversation generation with
  internal and external memory.
\newblock In {\em {AAAI}}, pages 730--739, 2018.

\end{thebibliography}

\end{document}